\documentclass[twocolumn]{htl-author}
\usepackage{float}
\usepackage{booktabs}
\usepackage{tabulary}
\usepackage{subfig}
\DOI{2024.0357}
\usepackage{lineno}
\newcolumntype{L}[1]{>{\raggedright\arraybackslash}p{#1}}
\newcolumntype{R}[1]{>{\raggedleft\arraybackslash}p{#1}}
\newcolumntype{C}[1]{>{\centering\arraybackslash}p{#1}}

\begin{document}
% \title{StraightTrack: A Trajectory-preserving Mixed Reality System for Navigated K-Wire Placement}
\title{StraightTrack: Towards Mixed Reality Navigation System for Percutaneous K-wire Insertion}
% \author{Anonymous}
\author{Han Zhang$^{*, 1}$, Benjamin~D.~Killeen$^{*, 1}$, Yu-Chun Ku$^{1}$, Lalithkumar Seenivasan$^{1}$, Yuxuan Zhao$^{1}$, Mingxu Liu$^{1}$, Yue Yang$^{1}$, Suxi Gu$^{1}$, Alejandro Martin-Gomez$^{1}$, Russell H. Taylor$^{1}$, Greg Osgood$^{2}$, and Mathias Unberath$^{1}$}
\address{$^{1}$Department of Computer Science, Johns Hopkins University, Baltimore, MD, USA\\
$^{2}$Department of Orthopaedic Surgery, Johns Hopkins Medicine, Baltimore, MD, USA\\
E-mail: hzhan206@jhu.edu\\
$^{*}$These authors contributed equally to this work.\\}

\historydate{Published in Healthcare Technology Letters; Received on xxx; Revised on xxx}

\abstract{In percutaneous pelvic trauma surgery, accurate placement of Kirschner wires (K-wires) is crucial to ensure effective fracture fixation and avoid complications due to breaching the cortical bone along an unsuitable trajectory. Surgical navigation via mixed reality (MR) can help achieve precise wire placement in a low-profile form factor. Current approaches in this domain are as yet unsuitable for real-world deployment because they fall short of guaranteeing accurate visual feedback due to uncontrolled bending of the wire. To ensure accurate feedback, we introduce StraightTrack, an MR navigation system designed for percutaneous wire placement in complex anatomy. StraightTrack features a marker body equipped with a rigid access cannula that mitigates wire bending due to interactions with soft tissue and a covered bony surface. Integrated with an Optical See-Through Head-Mounted Display (OST HMD) capable of tracking the cannula body, StraightTrack offers real-time 3D visualization and guidance without external trackers, which are prone to losing line-of-sight. In phantom experiments with two experienced orthopedic surgeons, StraightTrack improves wire placement accuracy, achieving the ideal trajectory within $5.26 \pm 2.29$\, mm and $2.88 \pm 1.49$\textdegree, compared to over 12.08\, mm and 4.07\textdegree for comparable methods. As MR navigation systems continue to mature, StraightTrack realizes their potential for internal fracture fixation and other percutaneous orthopedic procedures.}
\maketitle

\section{Introduction}

\begin{figure*}[b]
    \centering
    \includegraphics[width=\linewidth]{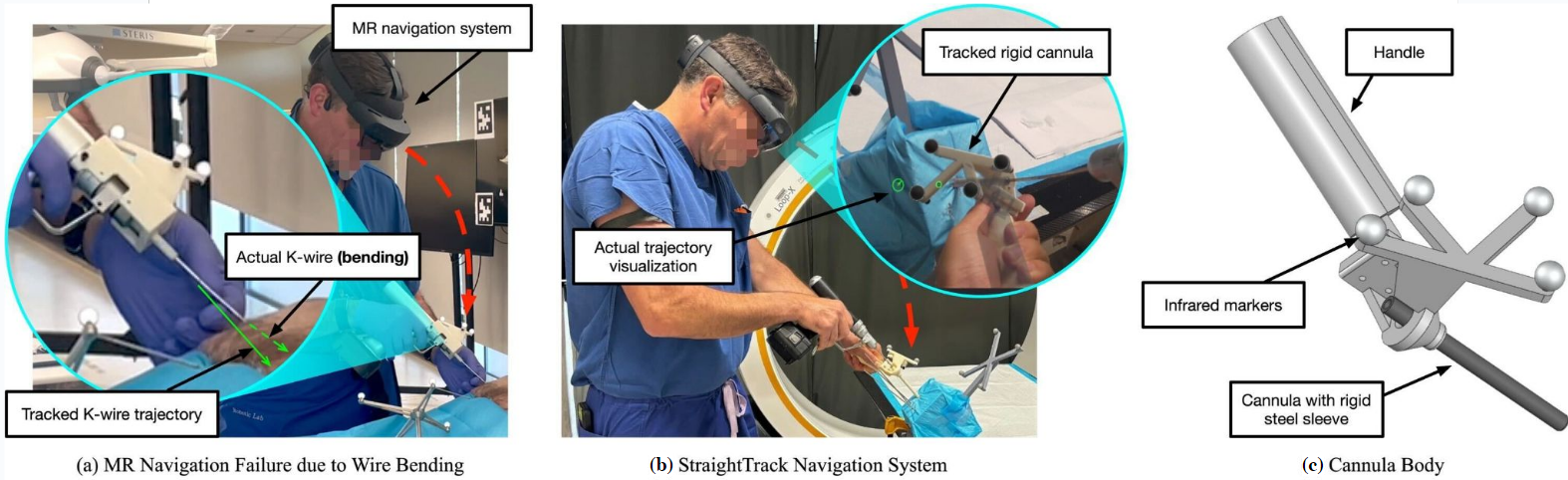}
    \caption{(a) A real-world failure mode of a conventional mixed reality (MR) navigation system for K-wire insertion, in which the K-wire bends in an uncontrolled manner. (b) Our StraightTrack navigation system visualizes the real-time K-wire trajectory error based on head-mounted tracking of the cannula body. The user aligns the cannula body by minimizing the radius of two circles, which visualize the entry- and endpoint error of the current trajectory. The hatch mark indicates the direction of the error. (c) The cannula body features a rigid steel sheath that ensures accurate tracking of the actual K-wire trajectory up to the cortical bone. This is necessary to overcome two real-world causes of K-wire bending, namely interaction with tough soft tissue and ``skating'' of the K-wire tip on hard bone.}
    \label{fig:system_overview}
\end{figure*}

The prevalence of pelvic ring fractures in the United States was 37 per 100,000 in 2023, with particularly high mortality rates observed among patients over 65 \cite{Davis2023May, Chong1997}. Effective treatment of these fractures often involves orthopedic surgery, such as pelvic fracture fixation, where the precise placement of screws is crucial for bone fixation and successful internal fracture management. To ensure this precision and avoid damage to vulnerable structures, accurate intra-operative guidance is essential. The cannulated screw system is widely used for this purpose, allowing surgeons to position a Kirschner wire (K-wire) along the desired trajectory before final screw placement, avoiding vulnerable structures and ensuring contact with the hard bone~\cite{Fahrenhorst-Jones2022Apr}. This system facilitates iterative adjustment of the K-wire trajectory based on X-ray image guidance, making it particularly valuable for minimally invasive internal fixation procedures in complex anatomical regions not only the pelvis, but also femoral neck, odontoid, and extremities \cite{Fahrenhorst-Jones2022Apr, Sen2005Feb, Wu2021Dec, Khoo2014Jul, Kim2015Nov, Yuan2022Dec, Galal2017Jul, Dickman1995Dec, Feger2022Jan, Acumed2024, Unberath2018Oct, Andress2018Apr}. These procedures are associated with smaller incisions, reduced pain, shorter recovery times, and overall better patient outcomes \cite{Wu2021Dec}. However, inserting K-wires in complex regions like the pelvis often necessitates multiple X-ray acquisitions and repeated attempts, even for experienced trauma surgeons \cite{Killeen2022Apr, Killeen2023Oct}. The difficulty arises from the inability of surgeons to directly visualize the bone, which increases the potential for human error during the procedure \cite{Killeen2024Jun, Killeen2022Apr}. Furthermore, this process can expose patients and clinicians to significant radiation, with each screw placement requiring between 20 to 100 fluoroscopic images \cite{Killeen2023Oct, Unberath2018Oct, DeSilva2018Jan}.

Together, these factors have motivated the development of computer-assisted approaches for cannulated screw placement that minimizes X-ray image acquisition while still providing effective intra-operative guidance. Recent methodologies~\cite{Fotouhi2019Jun, Killeen2023Jul, Wolf2023Aug, Liebmann2019Jul, Li2024Apr} employ 2 to 4 initial X-ray acquisitions to establish a desired trajectory, which is then visualized in 3D using Mixed Reality (MR) via an Optical See-Through Head-Mounted Display (OST HMD). MR guidance is particularly advantageous for these procedures as it maintains the dexterity and tactile feedback of freehand drilling while eliminating the need for bulky and expensive robotic systems, such as the Globus Medical ExcelsiusGPS Robotic Navigation Platform, costing approximately \$1.5 million USD \cite{Condon2024Feb}. Modern HMDs can serve as tracking devices, ensuring the accuracy of a fully navigated approach without requiring external optical trackers, which are difficult to position to maintain a clear line-of-sight~\cite{MartinGomez2020}. Despite the advancements in trajectory planning and the use of MR for guidance to reduce navigation errors during K-wire placement, challenges remain. K-wires often \textbf{bend unintentionally} due to the force exerted on them while maintaining a constant insertion angle on an irregularly shaped bone surface. This bending means that real-time tracking of the surgical drill's pose may not align with the K-wire's actual trajectory, as illustrated in Fig.~\ref{fig:system_overview}.a. There is an unmet clinical need for a surgical guiding system that ensures effective and accurate K-wire insertion in complex anatomy. To the best of our knowledge, no previous studies on tracking and navigation systems have investigated K-wire bending during insertion, nor have they addressed limiting K-wire bending or tracking it to compensate for this.  

In this paper, we introduce the StraightTrack, an MR system designed for orthopedic guiding wire placement that prevents K-wire bending in complex anatomy (Fig.~\ref{fig:system_overview}.b, \ref{fig:system_overview}.c). StraightTrack consists of a rigid access cannula equipped with a marker body that mitigates wire bending despite soft tissue and bone contact. Integrated with an OST HMD with on-device optical tracking, our system offers real-time surgical tracking and guidance, including a user interaction paradigm for mitigating difficulties in spatial navigation and reducing errors in entry point perception. Finally, we evaluate StraightTrack for traditional K-wire placement tasks in phantoms with two experienced orthopedic surgeons. Our results demonstrate that the proposed system effectively addresses the K-wire bending problem, reducing wire displacement errors in translation and rotation at the entry, mid, and endpoint, thereby improving the overall accuracy of K-wire placement.

\section{Methodology}
StraightTrack consists of two main components: a cannula body for maintaining desired trajectories and an OST-HMD (i.e., Microsoft HoloLens 2) for intra-operative tracking and navigation. For instrument tracking, we use the Time-of-Flight (ToF) depth sensor on the HMD to track retro-reflective spheres mounted on the instrumented and patient in a predefined layout\cite{10021890}. We utilize the STTAR package~\cite{martin2023sttar} to achieve marker detection and tracking. In addition to the Kalman filter integrated within the STTAR package, we introduced smooth interpolation between time frames to improve tracking stability and reduce jitter. We implemented the user interface and navigation using Unity3D and the Mixed Reality Toolkit 2~\cite{polar-kev2024May}.

\subsection{Cannula Body Design}

To prevent bending while enabling tracking during K-wire insertion, we designed a cannula body (Fig.~\ref{fig:system_overview}.c). The cannula body is composed of three primary components: the handle, marker mount, and wire sleeve. The marker mount is designed with tilted angles to maintain a clear line-of-sight when the surgeon holds the cannula body, ensuring that the retro-reflective markers are visible on different planes during positioning. In our experiment, the cannula body was prototyped using a Stratasys F170 Fused Deposition Modeling (FDM) 3D printer with Acrylonitrile Butadiene Styrene (ABS) material. The cannula body was fabricated in three separate parts, which were subsequently assembled using screws and nuts to enhance its strength. To prevent deformation of the plastic sleeve during wire positioning in tissue, a hollow steel shaft was inserted inside the plastic sleeve. 

To accurately determine the shaft axis of the cannula body and minimize errors introduced during the manufacturing phase, we conducted multiple pivot calibrations. The pivot calibration was performed by using Polaris NDI (Northern Digital Incorporated, Ontario, Canada). During each pivot calibration, a K-wire was inserted and locked in place. The K-wire was extended toward the cannula body tip in increments of 1 cm. In total, we performed six pivot calibrations, extending the K-wire by 6 cm. A regression analysis was conducted to determine the shaft axis. The pivot calibration resulted in a mean 3D RMS error of 0.66 mm and a mean error of 0.61 mm.

\begin{figure*}[!t]
    \centering
    \includegraphics[width=\linewidth]{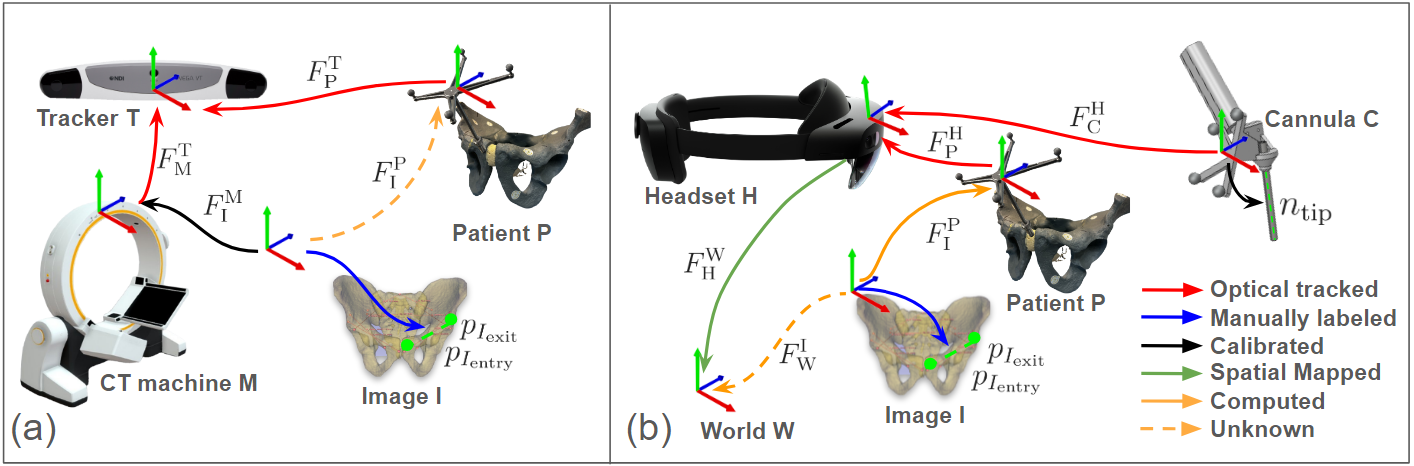}
    \caption{The transformation chain of the system. (a) Pre-operative trajectory planning phase: The trajectory points $\mathbf{P}_{\rm l_{entry}}$ and $\mathbf{P}_{\rm l_{exit}}$(shown in green points) were annotated from a CT image (shown with a blue arrow). The CT registration $F^{\text{P}}_{\text{I}}$ (shown with a yellow dotted arrow) is computed by using an optical tracker that tracks both the CT machine $F^{\text{T}}_{\text{M}}$ and the patient $F^{\text{T}}_{\text{P}}$ (shown with a red arrow). (b) Intra-operative guidance phase: The headset tracks both the patient $F^{\text{H}}_{\text{P}}$ and the cannula $F^{\text{H}}_{\text{C}}$(shown with a red arrow), and provides navigation to align $\mathbf{n}_{tip}$ with desired trajectory(shown in green dotted line) in world coordinate. The annotated entry and exit points in the world coordinate are computed using $F^{\text{I}}_{\text{W}}$(shown in a dotted orange arrow).}
    \label{fig:workflow}
\end{figure*}

\subsection{System Workflow}
The overall system workflow consists of pre-operative trajectory planning and intra-operative guidance. The overview of the transformation chain(Fig.~\ref{fig:workflow}). To provide effective K-wire placement guidance, the main goal of our system is to align the cannula body shaft axis $\mathbf{n}_{\rm tip}$ to trajectory axis $\mathbf{n}_{\rm trajectory}$ in world coordinate, as shown in Equation~\ref{equation8}.

\begin{equation}\label{equation8}
\begin{aligned}
F^{\text{W}}_{\text{H}} \cdot F^{\text{H}}_{\text{C}} \cdot \mathbf{n}_{\text{tip}} &= F^{\text{W}}_{\text{I}} \cdot \mathbf{n}_{{I_{trajectory}}} \\
\mathbf{n}_{{I_{trajectory}}} &= \mathbf{p}_{{I_{entry}}} - \mathbf{p}_{I_{exit}}
\end{aligned}
\end{equation}
\noindent

where $F^{\text{w}}_{\text{H}}$ is the transformation from World to HoloLens, $F^{\text{H}}_{\text{C}}$ is the transformation from HoloLens to Cannula body, ${n}_{tip}$ is the cannula body shaft axis in cannula body coordinate, $F^{\text{W}}_{\text{I}}$ is the transformation from World to CT image, and ${n}_{{I_{trajectory}}}$ is the trajectory axis in CT image coordinate, determined by the trajectory points $p_{\text{I}_{\text{entry}}}$ and $p_{\text{I}_{\text{exit}}}$ in CT image coordinates.
\newline

\textit{Pre-operative trajectory planning}: In the pre-operative trajectory planning phase (Fig.~\ref{fig:workflow}.a), the primary objectives are to annotate the ideal trajectory in the patient's CT image, denoted as $\mathbf{p}{I_{\text{entry}}}$ and $\mathbf{p}{I_{\text{exit}}}$, and to perform the registration of the CT image to the patient coordinate system $F^{\text{P}}_{\text{I}}$. The trajectory is either manually labeled in the CT image or determined using computer-assisted methods. 

Our system utilizes the Brainlab Loop-X in conjunction with the BrainLab Curve for CT imaging. The registration process involves an external optical tracker, as represented in Equation~\ref{equation1}.

\begin{equation}\label{equation1}
F^{\text{P}}_{\text{I}} = (F^{\text{T}}_{\text{P}})^{-1} \cdot F^{\text{T}}_{\text{M}} \cdot F^{\text{M}}_{\text{I}}
\end{equation}
\noindent
where $F^{\text{T}}_{\text{P}}$ represents the transformation from the Optical Tracker to the Patient array, $F^{\text{T}}_{\text{M}}$ is the transformation from the Optical Tracker to the CT machine, and $F^{\text{M}}_{\text{I}}$ denotes the transformation from the CT machine to the CT image coordinate system.

To achieve the tracking of the CT machine, represented by $F^{\text{T}}_{\text{M}}$, seven retro-reflective markers are placed on the CT machine gantry, enabling detection by the optical tracker. The pose of the CT machine is estimated using paired-point registration. The transformation $F^{\text{M}}_{\text{I}}$ is pre-calibrated by using a CT scan of an object embedded with both BB markers and optical markers arranged in a known configuration, with the registration performed using the paired-point registration.

With $F^{\text{P}}_{\text{I}}$ is determined and the ideal trajectory entry and exit points $\mathbf{p}{I_{\text{entry}}}$ and $\mathbf{p}{I_{\text{exit}}}$ are annotated, the pre-operative planning phase is completed.

\textit{Intra-operative guidance}: In the intra-operative guidance phases, the system provides tracking and visual navigation for K-wire placement using the HMD, as shown in Equation~\ref{equation2}.

\begin{equation}\label{equation2}
\begin{aligned}
F^{\text{W}}_{\text{H}} \cdot F^{\text{H}}_{\text{C}} \cdot \mathbf{n}_{\text{tip}} &= F^{\text{W}}_{\text{H}} \cdot 
F^{\text{H}}_{\text{P}} \cdot F^{\text{P}}_{\text{I}} \cdot \mathbf{n}_{{I_{trajectory}}} \\
\mathbf{n}_{{I_{trajectory}}} &= \mathbf{p}_{{I_{entry}}} - \mathbf{p}_{I_{exit}}
\end{aligned}
\end{equation}
\noindent
The transformation from world to HoloLens $F^{\text{w}}_{\text{H}}$ is estimated by the SLAM-based tracking system of the HMD. The pose of the patient anatomy $F^{\text{H}}_{\text{C}}$ and cannula body$F^{\text{H}}_{\text{P}}$ are tracked by HMD. CT registration $F^{\text{P}}_{\text{I}}$ and the annotated entry and exit point $\mathbf{p}_{I_{entry}}$ $\mathbf{p}_{I_{exit}}$ are obtained during pre-operative trajectory planning phase. To minimize errors introduced by SLAM drafting in the HMD, the system is designed to disable navigation if either the tool or patient momentarily loses track.

\begin{figure}[!t]
    \centering
    \includegraphics[width=\linewidth]{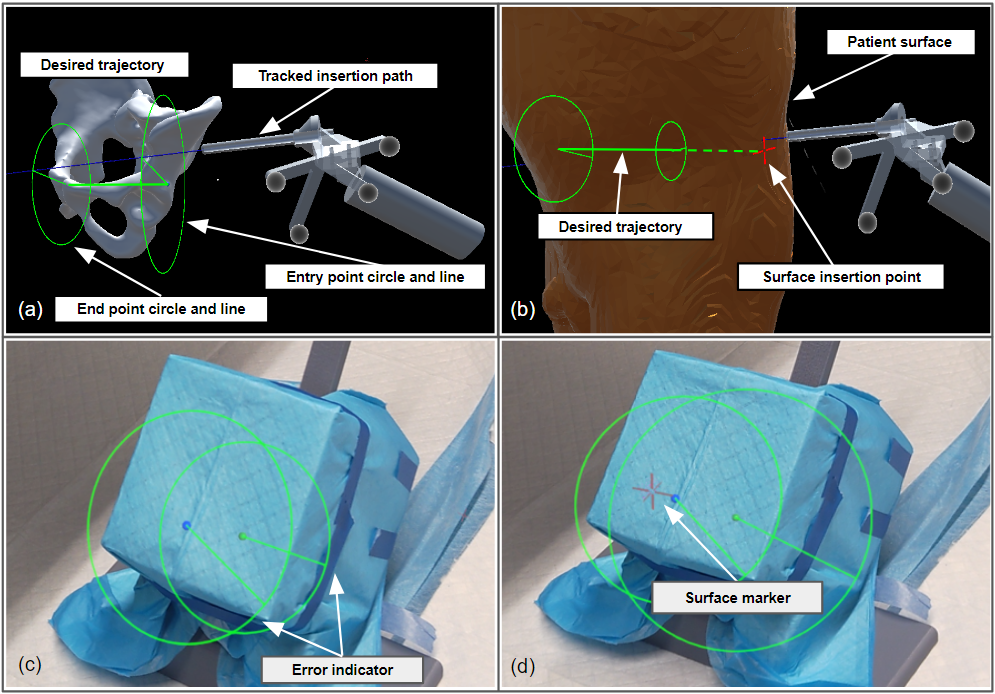}
    \caption{Navigation interface features for K-wire placement: (a) The error indicator: the user aligns the tracked insertion path(shown in blue line) with the desired trajectory(shown in green line). The circle and lines located at the entry point and end point serve as visual cues. Users need to manipulate the cannula body to minimize the size of green circles and lines. (b) The surface marker: the user positions the cannula body tip to the red cross mark, which is the insertion point on the patient's surface. The insertion point was generated by finding the closest point on the patient's surface to the desired trajectory. (c) The user perspective of error indicator in a phantom study. (d) The user perspective of surface marker and error indicator in a phantom study.}
    \label{fig:navigationinfo}
\end{figure}

\subsection{Navigation Interface}
In discussions with orthopedic surgeons, several key challenges in AR navigation for percutaneous surgery were identified and addressed in our navigation system. These challenges include:

\begin{itemize}
    \item \textbf{Spatial Alignment:} 
    Positioning surgical instruments in a 3D space demands precise alignment across six degrees of freedom. Relying solely on a surgeon's natural hand-eye coordination and depth perception requires constant adjustment and transformation between different views. This complexity makes it challenging to align the instruments accurately with both the skin entry point and the desired trajectory within the body during percutaneous surgery.

    \item \textbf{Incision Point Identification:} 
    Percutaneous surgery does not provide a direct view of the desired trajectory on the bone within the body. Misalignment of the skin incision with the desired trajectory can cause the K-wire to pass through soft tissues at an unfavorable angle. Further adjustment may bend the K-wire as it encounters resistance from soft tissues and bone, increasing the risk of deviation from the planned trajectory.
    
    \item \textbf{Clear Surgical Field:} 
    Excessive visual information on navigation interfaces may increase the surgeon's cognitive load and obstruct the surgical site. The proper navigation interface should provide essential guidance cues without causing unnecessary distraction or occlusion. 
    
\end{itemize}

To address these spatial alignment challenges during human-in-the-loop K-wire placement, we incorporated two features to provide real-time visual guidance: an Error Indicator and a Surface Marker (Fig.~\ref{fig:navigationinfo}). These design elements were inspired and refined through feedback from surgeons.

\begin{itemize}
    \item \textit{Error Indicator:} To ensure the correct spatial alignment of the K-wire and have a clear surgical site view, our system generates visual cues in the form of dotted circles and lines at the designated entry and exit points. This approach is similar to the method described by \cite{Jiang2023Sep}, but with modifications to eliminate unnecessary visual elements that may cause distraction or occlusion. The HMD simultaneously tracks the pose of the cannula body and the patient reference array, estimating the error by comparing the tracked insertion path to the desired trajectory. The radius of each dotted circle represents the deviation between the current and desired paths, while the direction indicated by the line suggests the necessary adjustment. The user aims to minimize the size of the circles to achieve optimal alignment.
        
    \item \textit{Surface Marker:} To accurately display the insertion point and prevent visual occlusion of the surgical site, the system visualizes the surface insertion point on the skin using the HMD's built-in depth camera. In our system, the HoloLens generates patient surface point clouds from its Articulated HAnd Tracking (AHAT) depth sensor. The position of the surface marker is calculated by identifying the closest point projected onto the desired trajectory. The orientation of the surface marker plane is determined by applying Singular Value Decomposition to the closest surrounding 500 points. During K-wire insertion, the user makes an incision at the surface marker and inserts the cannula at this location.
    
\end{itemize}

During the proposed wire placement workflow, desired trajectories are obtained via registration with a pre-operative plan or using an intra-operative workflow such as \cite{Killeen2023Jul, Fotouhi2019Jun}. The surgeon makes an initial incision in the soft tissue at the surface marker point and then inserts the cannula body until it reaches the bone surface. If necessary, the system can also function as a trocar for sharp instruments. Once in contact with the bone surface, the cannula body is aligned to the desired trajectory based on the error indicators. A sharp edge on the cannula allows it to grip the bone surface, preventing the K-wire from "skating" out of alignment before gaining purchase. Finally, the surgeon inserts the K-wire through the wire sleeve and begins the drilling process. 

\section{Experiments and Result}

\begin{figure}[!b]
    \centering
    \includegraphics[width=\linewidth]{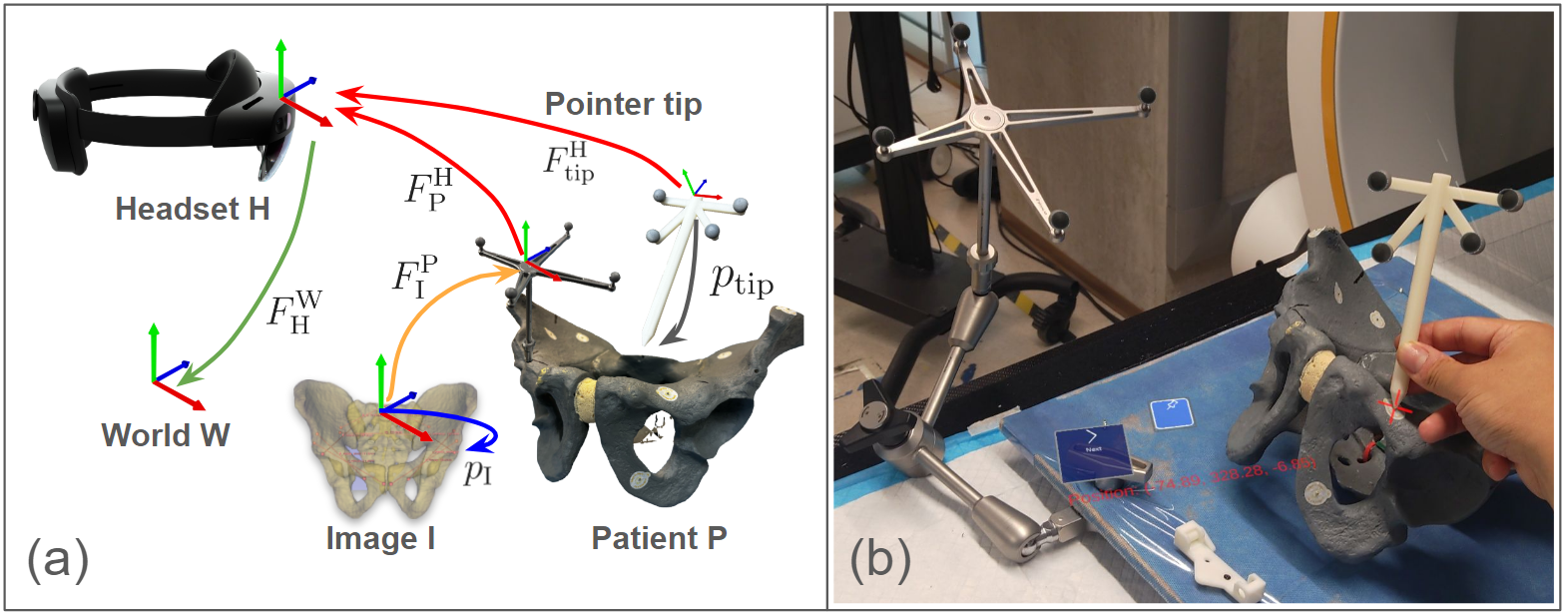}
    \caption{System end-to-end accuracy evaluation: (a) The transformation chain used in the study, consisting of the World (T), HoloLens (H), Patient (P), CT Image (I), and Pointer (tip). 'F' represents the transformation from one element (arrowback) to another (arrowhead). The transformation was obtained using optical tracking (shown in red), manual labeling (shown in blue), pre-computed values (shown in yellow), or spatial mapping (shown in green). (b) A test user points the tooltip (shown as a red cross mark) at a landmark on a pelvic phantom, with the printed tip position shown nearby.}
    \label{fig:end-to-end}
\end{figure}

We evaluate our fully integrated system with the surgeon in the loop. First, we validate the end-to-end tracking error after the pre-operative planning phase, using a touch-point evaluation. To evaluate the full system, we conduct a phantom study with experienced orthopedic surgeons performing a K-wire insertion procedure.

\subsection{Quantitative Assessment of End-to-End System Accuracy}
To quantitatively evaluate the overall registration and navigation accuracy of our MR system without human-in-the-loop, we perform an end-to-end system accuracy assessment (Fig.~\ref{fig:end-to-end}) using the method described below.

\subsubsection{Experimental Setup}
We place seven 1.5 mm metal bearing balls (BBs) landmarks on a pelvic phantom target. Following the proposed preoperative trajectory planning workflow (Fig.~\ref{fig:workflow}.a), we obtained a CT image and manually annotated their positions $\mathbf{p}_{\rm I}$. Using Equation~\ref{90}, we obtained CT registration $F^{\text{P}}_{\text{I}}$ as well as landmark position $\mathbf{p}_{\text{I}}$ and send them to the HMD: 

\begin{equation}\label{90}
F^{\text{P}}_{\text{I}} = (F^{\text{T}}_{\text{P}})^{-1} \cdot F^{\text{T}}_{\text{M}} \cdot F^{\text{M}}_{\text{I}}
\end{equation}
\noindent
To measure the end-to-end system accuracy, we use a 3D-printed pointer to sequentially touch the annotated landmarks. The pointer, with a 1.5 mm diameter sphere-shaped groove tooltip, ensures repeatable and precise placement on the metal BBs. To obtain the tip position $\mathbf{p}_{\rm tip}$, we performed a pivot calibration using the Polaris NDI. Since the pointer marker array has the same layout as the cannula body, the tracking error observed in the pointer marker can be correlated to the tracking error in the cannula body. The pointer is tracked by the HoloLens to locate the landmarks (Equation~\ref{10}).

\begin{equation}\label{10}
F^{\text{w}}_{\text{H}} \cdot F^{\text{H}}_{\text{tip}} \cdot \mathbf{p}_{\text{tip}} = F^{\text{W}}_{\text{H}} \cdot 
F^{\text{H}}_{\text{P}} \cdot F^{\text{P}}_{\text{I}} \cdot \mathbf{p}_{I} 
\end{equation}

\noindent By comparing the landmark positions computed in the pelvic phantom target coordinates with the landmark position obtained using the pointer tip, we can assess the accuracy of our MR solution. The end-to-end error, $\mathbf{\epsilon_{system}}$, can then be computed as:

\begin{equation}\label{11}
\mathbf{\epsilon _{system}} = \left |{F^{\text{W}}_{\text{H}} \cdot F^{\text{H}}_{\text{tip}} \cdot \mathbf{p}_{\text{tip}} - F^{\text{W}}_{\text{H}} \cdot 
F^{\text{H}}_{\text{P}} \cdot F^{\text{P}}_{\text{I}} \cdot \mathbf{p}_{I}} \right |
\end{equation}

This error can be broken down into several components: HMD SLAM mapping ($\epsilon^{\text{w}}_{\text{H}}$), HMD optical tracking ($\epsilon^{\text{H}}_{\text{tip}}$, $\epsilon^{\text{H}}_{\text{P}}$), instrument pivot calibration ($\epsilon_{\text{tip}}$), and CT registration ($\epsilon^{\text{P}}_{\text{I}}$, $\epsilon_{\text{I}}$).

\subsubsection{Quantitative Results}
For the pointer pivot calibration, we observed a 3D RMS error of 0.58 mm and a mean error of 0.48 mm. For the end-to-end accuracy evaluation of the system, \textbf{the average system error was 2.89 mm $\pm$ 0.97 mm}. This result represents the overall error in reaching a point labeled in the CT volume in real space using our MR navigation system, excluding human-in-the-loop error.

\begin{figure*}[!t]
    \centering
    \includegraphics[width=\linewidth]{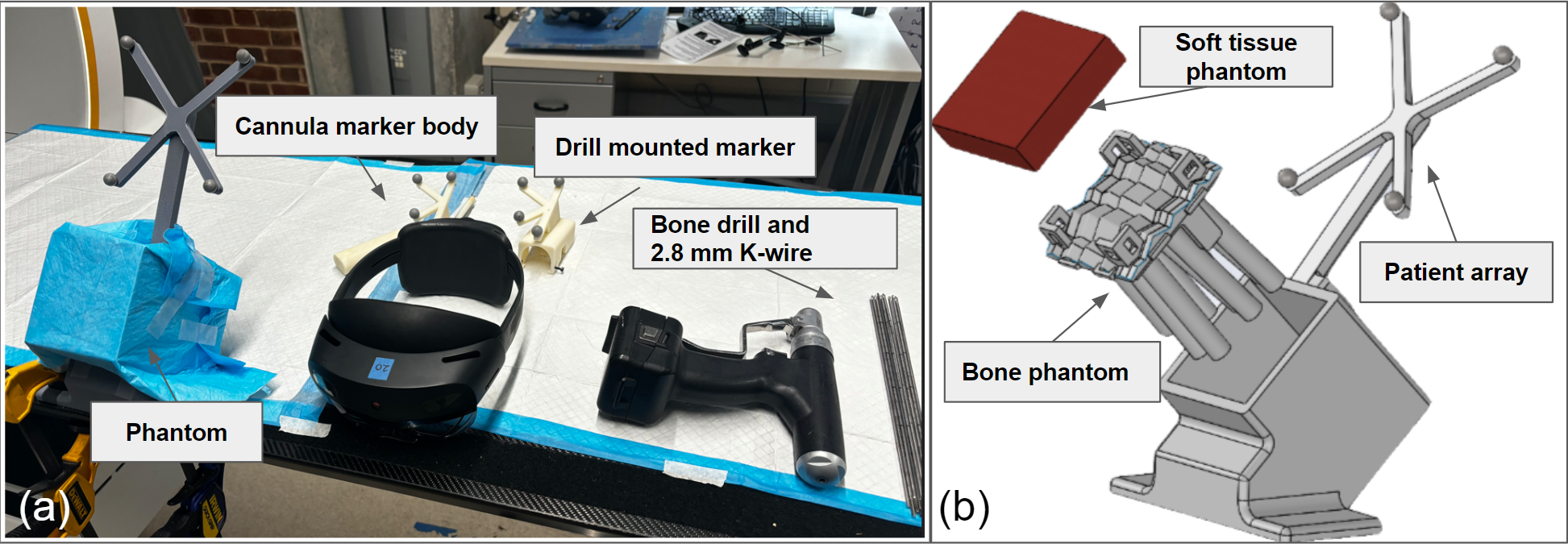}
    \caption{Phantom study: (a) Experimental setup for inserting 2.8 mm k-wires under MR guidance with different marker mounts.(b) Components of the phantom, include a silicone rubber soft tissue phantom, a 3D-printed bone structure phantom, and a holder with a patient array marker.}
    \label{fig:experiment}
\end{figure*}

\subsection{Phantom study}
To investigate the end-to-end, human-in-the-loop error of the StraightTrack system, we conducted a controlled phantom study focused on K-wire insertion in the presence of soft tissue and an irregular bone surface, as is the case in navigated fracture fixation. The experimental setup for our phantom study is shown in Fig.~\ref{fig:experiment}.a. During the experiment, the participant used a HoloLens HMD to place wires along specified trajectories, employing various guidance interfaces and marker mount methods. The cannula-mounted interface is our StraightTrack system.

\subsubsection{Phantom Preparation}
We manufactured five phantom models to replicate the traditional K-wire placement procedure in patients, as illustrated in Fig.~\ref{fig:experiment}.b. Each phantom consists of a 3D-printed structure with nine cylindrical shapes, a cast soft tissue layer, and a holder equipped with a retro-reflective marker array. The bone was fabricated using ABS material (Stratasys F170 FDM 3D printer) with 60\% material infill to better simulate the properties of human bone material and the drilling experience\cite{Kiel-Jamrozik2022Jun}. To better replicate the difficulty of drilling on complex anatomy, such as the pelvis, each 100 mm cylinder was oriented in a random direction. The angle of orientation was less than 15\textdegree relative to the perpendicular. Additionally, each cylinder featured an uneven shape on the contact surface to simulate an irregular bone surface. The structure was then filled with 20 mm of silicone rubber gel (Ecoflex 00-35, Smooth-On, Inc. , PA) to mimic human soft tissue and was held with laser-cut acrylic boards. The structure and the holder were connected with screws to maintain a rigid transformation.

% \begin{figure}[!t]
%     \centering
%     \includegraphics[width=\linewidth]{pic/UI_three.png}
%     \caption{Three navigation interfaces in the phantom study: (a) Non-tracked: An extended line connecting the desired entry and exit points. (b) Error Indicator: The cross point of dotted circles and lines represents the error between the current insertion path and the desired entry and exit points. (c) Error Indicator with Surface Marker: A cross mark on the soft tissue indicates the surface incision point, in addition to the error indicator.}
%     \label{fig:UI_three}
% \end{figure}

\subsubsection{Study Design}
\begin{figure}[!t]
    \centering
    \includegraphics[width=\linewidth]{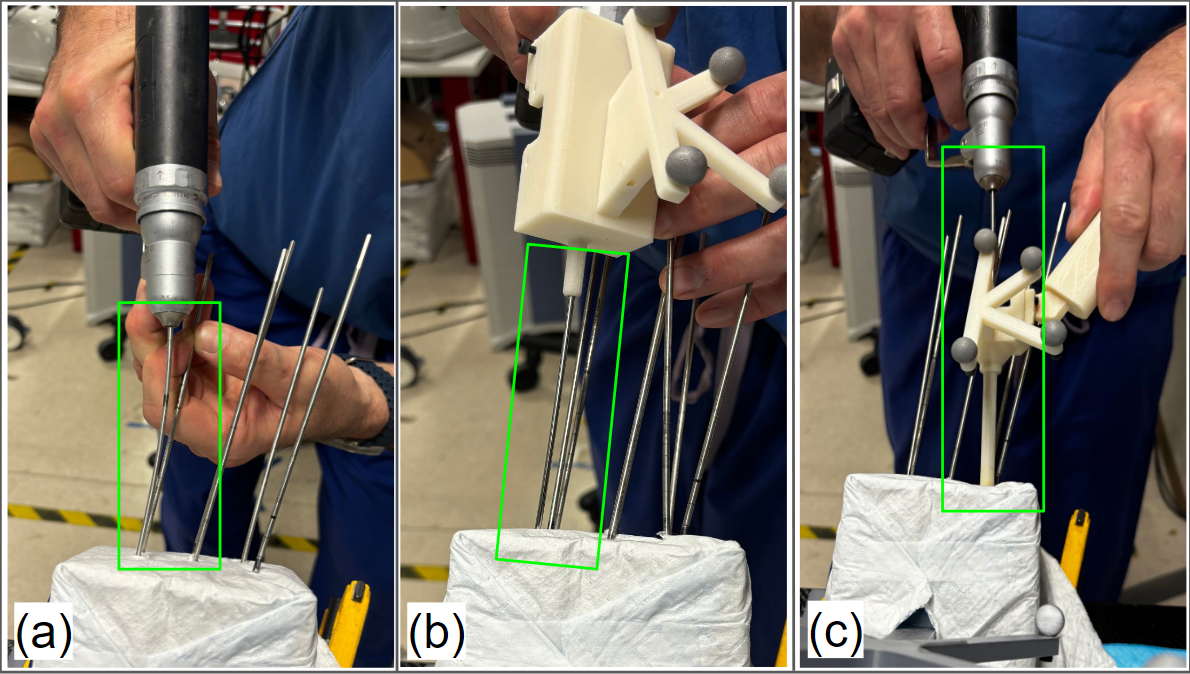}
    \caption{A surgeon drilling a K-wire using different marker mount methods during the phantom experiment, with wire bending observed. (a) Non-tracked method. (b) Drill-mounted marker method. (c) Cannula body method.}
    \label{fig:wire_in_three_method}
\end{figure}

In our phantom study, we invited two experienced orthopedic surgeons ((with over 10 years of experience) to place 2.8 mm K-wires. We compared three different marker mount methods (non-tracked, drill-mounted marker, and cannula body), as shown in Fig.~\ref{fig:wire_in_three_method}, and two navigation interfaces (error indicator with and without the surface marker). For the non-tracked method, we did not track the K-wire; instead, we displayed a line connecting the entry and exit points as the navigation interface. In total, we examined six different methods resulting from the combinations of mounting methods and navigation interfaces.

\begin{table*}[!t]
    % \centering
    \begin{tabular}{lrrrr}
        \toprule
        \textbf{Navigation Target} & \textbf{Entry Point Error(mm)} & \textbf{Mid Point Error (mm)} & \textbf{End Point Error (mm)} & \textbf{Rotation Error (\textdegree)}\\
        \midrule
        Non-tracked & 8.91 $\pm$ 5.35 & 6.75 $\pm$ 3.84 & 6.52 $\pm$ 2.17& 3.95 $\pm$ 1.86\\
        Drill & 12.08 $\pm$ 5.29 & 8.69 $\pm$ 4.56 & 6.66 $\pm$ 2.94& 4.07 $\pm$ 2.19\\
        \textbf{Cannula (ours)} & \textbf{5.26 $\pm$ 2.29}& \textbf{3.85 $\pm$ 1.10} & \textbf{3.57 $\pm$ 1.52}& \textbf{2.88 $\pm$ 1.49}\\
        \bottomrule
        \end{tabular}
    \vspace{5pt}
    \caption{Final K-wire placement error for three MR navigation systems, including visualization only (non-tracked), a drill-mounted tracked system, and our cannula-mounted StraightTrack system.}
    \label{tab:comparison}
\end{table*}

\begin{table*}[!t]
    % \centering
    \begin{tabular}{lrrrr}
        \toprule
        \textbf{Navigation Target} & \textbf{Entry Point Error(mm)} & \textbf{Mid Point Error (mm)} & \textbf{End Point Error (mm)} & \textbf{Rotation Error (\textdegree)}\\
        \midrule
        Non-tracked & 10.62 $\pm$ 5.09& 7.52 $\pm$ 3.21& 5.21 $\pm$ 1.91& 4.31$\pm$ 2.14\\
        Drill & 9.76 $\pm$ 5.43& 7.16 $\pm$ 3.48& 6.02 $\pm$ 2.36& 4.22 $\pm$ 3.03\\
        \textbf{Cannula (ours)} & \textbf{8.68 $\pm$ 5.54}& \textbf{6.39 $\pm$ 3.45}& \textbf{4.72 $\pm$ 1.79}& \textbf{3.35 $\pm$ 2.57}\\
        \bottomrule
        \end{tabular}
    \vspace{5pt}
    \caption{Final K-wire placement error for three MR navigation systems when using HoloLens depth sensor to provide a surface marker for initial insertion.}
    \label{tab:comparison-marker}
\end{table*}

\begin{table}[]
    % \centering
    \begin{tabular}{lrr}
    \toprule
     $p$-value vs: & Non-tracked & Drill mount\\
    \midrule
    Entry Point Error & \textbf{0.0411}& \textbf{0.0005}\\
    Midpoint Error & \textbf{0.0199}& \textbf{0.0017}\\
    Endpoint Error & \textbf{0.0008}& \textbf{0.0038}\\
    Rotation Error & 0.1376& 0.1336\\
    \bottomrule
    \end{tabular}
    \vspace{5pt}
    \caption{Statistical significance of StraightTrack.}
    \label{tab:p-value}
\end{table}

\begin{table}[]
    % \centering
    \begin{tabular}{lrr}
    \toprule
     $p$-value vs: & Non-tracked & Drill mount\\
    \midrule
    Entry Point Error & 0.3808& 0.6356\\
    Midpoint Error & 0.4138& 0.591\\
    Endpoint Error & 0.5281& 0.1438\\
    Rotation Error & 0.3285& 0.4573\\
    \bottomrule
    \end{tabular}
    \vspace{5pt}
    \caption{Statistical significance of StraightTrack with the surface marker.}
    \label{tab:p-value-marker}
\end{table}

During the preoperative planning phase, we collected CT volumes from the phantoms and planned the target trajectory paths for each cylinder. By computing the spatial relationship between the planned trajectory points and the patient marker array using Equation~\ref{equation1}, the pre-planned trajectories were made visible to the surgeon through the HMD. The surgeon performed eye-tracking calibration before starting the procedure. Before the actual experiment, the surgeon practiced multiple K-wire insertions in a test phantom to become familiar with the setup and flatten the learning effect.

Following this familiarization, each surgeon performed K-wire insertions under different MR guidance methods. Each method was tested in six drilling trials, resulting in a total of 36 trajectories. The order of navigation interfaces and mounting markers was randomized to minimize the learning effect imbalance. Post-CT scans of each phantom were acquired to measure the difference between planned and actual trajectories. We report the final error of K-wire placements relative to the ideal, annotated trajectory in terms of the translational error at the entry, mid, and endpoint of the corridor, as well as the angle deviation between them.

\subsubsection{Study Results}
\begin{figure}[!t]
    \centering
    \includegraphics[width=\linewidth]{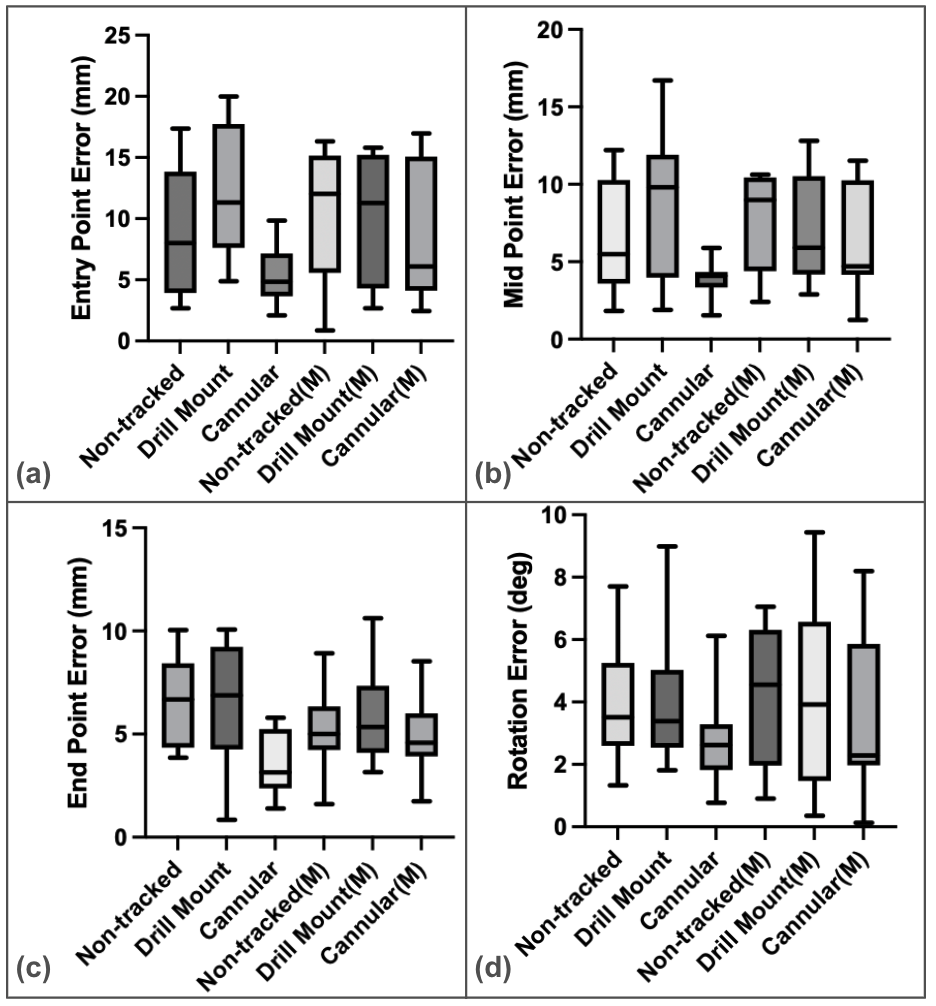}
    \caption{Phantom study results of K-wire insertion with different navigation interfaces and marker mount methods (M represents the use of surface marker). The entry(a), mid(b), and end(c) point displacement errors and rotation(d) errors between the inserted and planned trajectories were measured.}
    \label{fig:result_all}
\end{figure}

A summary of the phantom study results is presented in Fig.~\ref{fig:result_all} and Table.~\ref{tab:comparison}. We conducted 12 trials for each combination of marker mount methods and navigation interfaces. Among all six combinations (Table.~\ref{tab:comparison}.a), the method utilizing the cannula body with an error indicator achieved the lowest rotation error ($2.88\pm 1.49$\textdegree), as well as the lowest entry, mid, and endpoint error of $5.26\pm 2.29$\, mm, $3.85 \pm 1.10$\, mm and $3.57 \pm 1.52$\, mm, respectively. As shown in Table~\ref{tab:p-value}, the cannula body method demonstrated statistically significant improvements in translation error compared to both the non-tracked method and the drill mount method ($p < 0.05$). A high error indicates a high likelihood of cortical breach, highlighting the value of StraightTrack's cannula body for mitigating errors during the human-in-the-loop workflow.

The same results were observed when using the surface marker to indicate the entry point for incisions, with the StraightTrack cannula system achieving a low rotation error of $3.35\pm 2.57$\textdegree, as well as low start, mid, and endpoint errors of $8.68 \pm 5.54$ mm, $6.39 \pm 3.45$ mm, and $4.72 \pm 1.79$\, mm, respectively (see Table~\ref{tab:comparison-marker}). However, the effect size in this initial study is not statistically significant when using the surface marker, as in Table~\ref{tab:p-value-marker}. Similarly, when comparing similar tracking strategies with and without the surface marker, we observe an inconsistent effect on the final placement, with the surface marker sometimes improving final accuracy (as in the drill-tracked solution) and sometimes reducing it (as in the non-tracked and cannula-tracked solutions, including StraightTrack). In practice, providing the surface marker is valuable for estimating the incision point, but these results suggest it may not add value when assisting with the actual placement.

\section{Discussion and Limitation}

Addressing wire bending is crucial for bringing MR navigation systems into percutaneous procedures. Our experimental results strongly suggest that the cannula body design improves the accuracy of wire placement significantly. In particular, during the phantom study, we observed significant bending of the K-wire with non-tracked and drill-mounted marker methods when adjusting the wire within the soft tissue, as shown in Fig.~\ref{fig:wire_in_three_method}. In contrast, the cannula body effectively reduces the deviation caused by irregular forces during drilling, thereby ensuring that the MR system’s visual guidance remains precise and reliable. Moreover, we observed that the bending issue caused the drill-mounted MR navigation to perform worse than the non-navigation method. We hypothesize that the bent wire misled the surgeon and resulted in unnecessary adjustments. This highlights the need to integrate both accurate navigation and physical stabilization to achieve precise wire placement, as improvement in only one aspect may not yield better results. Although the statistical significance is not robust enough for definitive conclusions, our results show trends that strongly suggest the benefits of the cannula body design. We plan to conduct larger-scale studies to validate these findings and further optimize the cannula body design, enhancing the reliability of percutaneous procedures.

Furthermore, results differentiating only navigation interfaces show that the introduction of surface markers actually decreased performance. This suggests that additional visual aids should be designed to avoid introducing new sources of error. We hypothesize that uncertainty introduced by the device's depth sensor may lead surgeons to incorrectly identify the starting point on the bone. This uncertainty may arise from the Hololens depth sensor's native accuracy in capturing the complex surface topography of soft tissue. Specifically, the ToF depth sensor exhibits inconsistent depth estimation influenced by lighting conditions and surface characteristics such as reflectivity, texture, color, and material \cite{Gu2021May, Gsaxner2023Apr}. This incorrect surface marker positioning results in the misalignment of the initial insertion angle of the K-wire in the soft tissue, leading to greater overall errors in the wire trajectory.

Although only two expert orthopedic surgeons participated in the preliminary phantom study, our main objective was to investigate whether K-wire bending leads to poor outcomes and to assess the feasibility of our system in addressing this issue. The inclusion of highly experienced (>10 years) surgeons helps minimize variability due to human factors. Despite the limited number of trials for each method, there is a clear trend indicating that the cannula body provides stable performance. These insights are crucial for refining and developing the next version of our system to ensure accurate intraoperative wire placement.

\section{Clinical Applicability}

Seamless integration into clinical workflows is crucial for the successful adoption of any surgical navigation system. Our proposed system requires the use of a Dynamic Reference Body (DRB) with reflective spheres attached to the patient’s body. While the DRB is commonly used in surgical navigation for accurate real-time tracking~\cite{Cho2012Nov, Taylor2003Oct, Taylor2016Jul}, it presents challenges such as ensuring stable fixation, maintaining sterility, and minimizing procedural interference. Further investigation is needed to evaluate the practicality of DRB attachment across different surgical scenarios and explore alternative tracking methods to optimize safety and procedural efficiency.

The accuracy of the StraightTrack system, as demonstrated in our phantom studies, shows an average translation error of less than 5.26 mm and a mean rotational error of 2.88\textdegree. In the context of percutaneous pelvic trauma surgery, such as the fixation of superior pubic ramus fractures—a common injury often associated with pelvic ring damage—the average corridor width of the superior pubic ramus is reported to be 8.2 $\pm$ 1.8 mm \cite{Altinayak2023Mar}. Given this anatomical context, our system's current accuracy is approaching, but not yet meeting, the precision required for such procedures. For more delicate orthopedic surgeries, particularly those involving proximity to critical anatomical structures, the existing error margins still require improvement. Consequently, future efforts will focus on enhancing the system's precision to better align with the requirements of complex surgical interventions. Moreover, additional validation in more realistic settings, such as cadaver studies, is necessary to obtain clinically relevant data for our system.

As an initial prototype, our system provides a lightweight, cost-effective, and comprehensive MR solution for guiding orthopedic wire placement. The system is designed to be modular, scalable, and integrate with current technologies, including a range of AR headsets. Although our prototype was developed on the HoloLens platform, from an engineering perspective, it is compatible with other AR headsets that incorporate time-of-flight depth sensors, such as the Magic Leap 2. As technology progresses, ensuring the StraightTrack system's adaptability across hardware platforms is crucial.

\section{Conclusions and Future Work}

In this paper, we introduced StraightTrack, an MR-based trajectory-preserving navigation system for K-wire placement. StraightTrack integrates spatial navigation using an OST-HMD and a hand-held cannula body. We established the end-to-end system navigation accuracy and conducted a preliminary phantom study with experienced orthopedic surgeons. The results indicate that our system effectively addresses the wire bending issue and achieves higher wire placement accuracy, with an average translation error of less than 5.26\, mm and a mean rotational error of 2.88\textdegree.

For future work, we plan to conduct a comprehensive user study involving orthopedic surgeons with varying levels of expertise to further evaluate the system's performance and gather subjective feedback for refining the design. Additionally, we plan to perform a cadaver study to validate the system's effectiveness in a more realistic surgical setting. We will also fabricate the prototype with higher-quality machining and a smaller radius to enable less invasive approaches. Our goal is to conduct translational research to bring our system into clinical use.

% \section{Author Contribution}
\section{Conflict of Interest Statement} 
The authors declared that they have no conflicts of interest
\section{Data Availability Statement}
The research data are not shared.
\bibliography{references}
\bibliographystyle{ieeetr}

\end{document}